\pdfoutput=1 % arXiv: force pdflatex
\documentclass{article} % For LaTeX2e
\usepackage{iclr2027_conference,times}

\usepackage{amsmath,amsfonts,bm}

\def\eqref#1{equation~\ref{#1}}
\def\1{\bm{1}}

\DeclareMathAlphabet{\mathsfit}{\encodingdefault}{\sfdefault}{m}{sl}
\SetMathAlphabet{\mathsfit}{bold}{\encodingdefault}{\sfdefault}{bx}{n}

\usepackage{url}
\usepackage{tcolorbox}
\tcbuselibrary{skins}
\usepackage{soul}
\usepackage{color}
\usepackage{graphicx}
\usepackage{tikz}
\usetikzlibrary{arrows.meta}
\usepackage{amsmath}
\usepackage{amssymb}
\usepackage{booktabs}
\usepackage{algorithm}
\usepackage{algpseudocode}
\usepackage{wrapfig} 
\usepackage{needspace}
\usepackage{placeins}
\usepackage{multirow}
\usepackage{enumitem}
\usepackage{mathtools}
\usepackage{tabularx}
\usepackage{colortbl}
\usepackage[normalem]{ulem} % Required by \reviewdel.
\usepackage{xcolor}
\usepackage{hyperref}

\algrenewcommand\algorithmicrequire{\textbf{Input:}}
\algrenewcommand\algorithmicensure{\textbf{Output:}}
\algrenewcommand\algorithmicindent{1.0em}

\newcommand{\layout}{x}            % layout (binary pixel grid)
\newcommand{\target}{y}            % target response
\newcommand{\context}{c}           % context (e.g., board layout, environmental conditions)
\newcommand{\simulator}{{\bm F_{\mathrm{sim}}}} % full-wave simulator
\newcommand{\surrogate}{{\bm F}_{\mathrm{surr}}} % neural approximation
\newcommand{\simloss}{\mathcal{L}_{\mathrm{sim}}}
\newcommand{\surrloss}{\mathcal{L}_{\mathrm{surr}}}
\newcommand{\surrlosscontinuous}{\widetilde{\mathcal{L}}_{\mathrm{surr}}}
\newcommand{\budget}{B}            % hard budget of simulator calls
\newcommand{\dist}{d}              % response distance metric

\newcommand{\numsim}{N_{\simulator}}    % simulator call count

\ifdefined\cleanreview
    \DeclareRobustCommand{\reviewcomment}[2]{}
    \DeclareRobustCommand{\reviewdel}[1]{}

\else
    \DeclareRobustCommand{\reviewcomment}[2]{%
    \textcolor{blue}{\small\noindent \textbf{[#1]} #2}}
    \DeclareRobustCommand{\reviewdel}[1]{{\color{blue}\hypersetup{citecolor=blue,urlcolor=blue,linkcolor=blue}\sout{#1}}}

\fi

\title{Simulator-Refined Diffusion for\\ Radio-Frequency Inverse Design}

\newcommand{\authorcell}[3]{%
    \begin{minipage}[t]{0.49\textwidth}
        \normalfont\raggedright
        {\bf\rule{0pt}{24pt}#1}\\
        #2\\
        \texttt{#3}%
    \end{minipage}}
\author{%
    \authorcell{Jinhao Liang}{University of Virginia}{njs4nu@virginia.edu}%
    \authorcell{Jacob K. Christopher}{University of Virginia}{csk4sr@virginia.edu}\\[14pt]
    \authorcell{Michael Frei}{Arena Physica}{michael.frei@arenaphysica.ai}%
    \authorcell{Tommaso Dreossi}{Arena Physica}{tommaso.dreossi@arenaphysica.ai}\\[14pt]
    \authorcell{Nando Fioretto}{University of Virginia}{fioretto@virginia.edu}%
}

\iclrfinalcopy % Uncomment for camera-ready version, but NOT for submission.
\begin{document}

\maketitle
% Remove the ICLR header banner ("Under review…" / "Published as…")
\lhead{}

\begin{abstract}
Diffusion models have shown potential in inverse design of printed circuit boards (PCBs), enabling the generation of layouts conditioned on target S-parameters. Despite this promise, applying diffusion models to PCB layout generation remains challenging due to their difficulty in meeting the quantitative electromagnetic specifications. A common approach is gradient-based guidance, which biases the diffusion sampling process with the gradient of an objective used for evaluation. However, full-wave electromagnetic simulators are accurate but expensive and typically non-differentiable, whereas differentiable surrogates are informative but not always reliable.
To address these limitations, this paper proposes \emph{Simulator-Refined Diffusion} (SRD), a novel combination of a low-fidelity differentiable surrogate and a high-fidelity non-differentiable simulator within the diffusion sampling process. Unlike standard zeroth-order optimization, which requires a great number of random perturbations, our approach uses the surrogate's gradient to propose the perturbation direction while the simulator then searches based on this direction to identify an effective design update.
Experimental results across different settings show that this method consistently outperforms current state-of-the-art methods, producing layouts whose simulated S-parameters match the target specifications up to \textit{21.2\% closer} for in-distribution targets and up to \textit{19.8\% closer} for out-of-distribution targets. 
\end{abstract}

\section{Introduction}
\label{sec:introduction}

Radio-frequency (RF) printed circuit board (PCB) layout design asks for a physical geometry whose electromagnetic response matches a given specification. These specifications are often expressed in terms of scattering parameters (S-parameters) across a frequency range, which describe how an RF signal is transmitted and reflected by the layout.
Producing the desired response is a fundamental inverse problem in electromagnetics with applications spanning metallic antennas, microwave filters, and RF passives~\citep{hassan2014topology,aage2017topology}. While the mapping from geometry to response is well-defined, the inverse mapping is ill-posed and non-unique. Even for a fixed board outline, ports, and substrate, small geometric changes can alter S-parameters across frequency, while each high-fidelity evaluation requires a full-wave solution of Maxwell's equations~\citep{aage2017topology,saha2026ktrail}. The resulting inverse problem is both high-dimensional and expensive to verify, making it a natural candidate for machine learning methods that can learn from past designs to propose new candidates.

\begin{figure}[!tb]
    \centering
    \includegraphics[width=\textwidth]{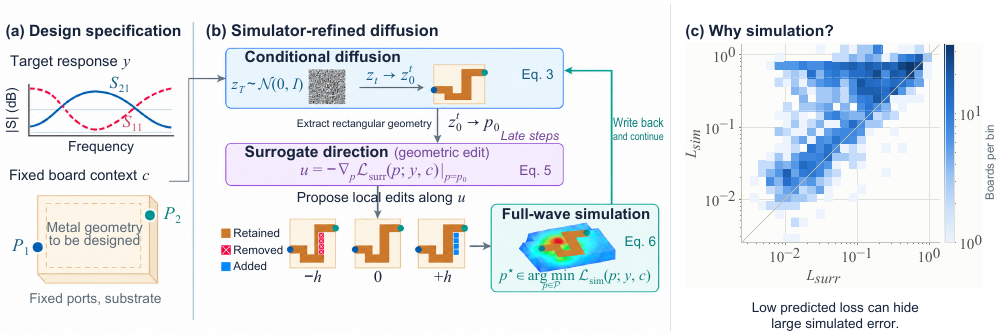}
    \vspace{-20pt}
    \caption{\textbf{Simulator-Refined Diffusion.}
    \textbf{(a)} Target S-parameters and board context condition generation.
    \textbf{(b)} At selected late denoising steps, surrogate gradients propose
    geometric edits (Eq.~\ref{eq:surrogate_direction}); the full-wave simulator
    selects the candidate to retain (Section~\ref{sec:propose_select}).
    Relative to the central unedited candidate, gold-colored pixels are retained, crossed red pixels are removed, and blue-colored pixels are added. 
    \textbf{(c)} Low correlation between the surrogate and the simulator downstream losses (RMSE) across tasks, observed especially in the low-loss regime and out of distribution regions.}
    \label{fig:overview}
\end{figure}

Conditional diffusion models offer a natural prior for this problem~\citep{ho2020denoising,11103838,dreossi2026inverse}. These models learn the conditional distribution of layouts given a target S-parameter response and available design metadata, so at generation time, they can produce diverse candidate layouts by sampling from this distribution. Yet conditioning alone does not ensure that a generated layout reproduces the target under a full-wave simulator~\citep{saha2026ktrail}. A common approach is to guide denoising with gradients from a differentiable surrogate~\citep{chung2023diffusion,ye2024tfg,zampini2026training}. In RF design, using a learned response surrogate makes this guidance inexpensive, but the final design then depends on surrogate accuracy along the sequence of edits used to approach the requested response~\citep{saha2026ktrail}.
Figure~\ref{fig:overview}(c) illustrates this mismatch where low predicted loss can coexist with large simulated error, and our experiments show that surrogate guidance alone attains negligible improvements over unguided diffusion (Table~\ref{tab:method_comparison}).

This paper starts from the hypothesis that a surrogate gradient can provide an informative direction for modifying the current design even when the appropriate displacement is uncertain. A full-wave simulator can then compare a small number of signed displacements without estimating a gradient in the full layout space. This separation converts simulator use from random exploration into targeted model selection. It also suggests when to spend the budget. Early in reverse diffusion, the inferred layout remains unstable and later denoising can overwrite an expensive edit. Simulator refinement is therefore concentrated near the end of the trajectory, when the predicted geometry is sufficiently resolved and the selected modification can persist to the final design.

We instantiate this principle in \emph{Simulator-Refined Diffusion} (SRD). At a selected denoising step, SRD extracts a structured geometric representation from the predicted clean sample and uses the surrogate to propose an edit direction. An initial full-wave simulator probe assesses whether a trial step along the proposed direction sufficiently improves the match between the simulated and target responses. The probe outcome determines whether the remaining simulation budget is allocated to evaluating additional step sizes along that direction or exploring alternative directions. SRD then writes the candidate with the lowest simulated loss back into the diffusion state before resuming denoising. The learned prior, surrogate, and simulator consequently serve complementary roles: the prior maintains plausible global structure, the surrogate focuses the search, and the simulator selects the edit that best matches the physical objective.

\noindent{\bf Contributions.} This paper makes three key contributions: {\bf (1)} it formulates RF inverse design as a budgeted inference problem in which total full-wave simulator calls are the primary resource; {\bf (2)} it introduces SRD, a structured directional-refinement procedure that uses surrogate gradients to propose edits and simulator evaluations to select effective design updates under an explicit call budget; and {\bf (3)} it evaluates this procedure on 250 real targets spanning 25 RF layout templates and 50 constructed out-of-distribution targets, achieving reductions in simulated S-parameter error of up to 21.2\% and 19.8\%, respectively, relative to the conditional diffusion baseline, \textit{nearly twice the margin of the next best approach}.
% \reviewcomment{Nando}{Can we claim the benchmark is a new contribution of this work? Also, please comment on results, how better we do w.r.t. SOTA.}

% \reviewcomment{Nando}{We should report here and add experiments about the arbitrary-layout repair and out-of-distribution evaluation promised under distribution shift.}

\section{Related work}
\label{sec:related_work}

% \reviewdel{Planned topics: electromagnetic and RF inverse design; guidance in diffusion models; inference-time search over diffusion samples; differentiable perturbed optimization; and surrogate-in-the-loop optimization.}

\noindent{\bf Electromagnetic and RF inverse design.}
Topology optimization has been used to design metallic antennas and microwave components under electromagnetic objectives~\citep{hassan2014topology,aage2017topology}. More recently, learned generative models have amortized part of this search by mapping desired responses to candidate layouts. \citet{11103838} synthesize EM structures from target S-parameters with a conditional diffusion model. In the RF setting, \citet{dreossi2026inverse} train a conditional diffusion model for the inverse design of PCB layouts and show that surrogate-based ranking can improve response matching. %A concurrent approach combines a conditional diffusion prior with black-box electromagnetic evaluations through derivative-free ensemble Kalman corrections~\citep{saha2026ktrail}. Our focus is complementary: 
Our approach extends these methods by using a differentiable surrogate to propose a structured geometric direction and reserves the full-wave simulator for selecting a signed displacement under a small call budget. 
\\[4pt]
\noindent{\bf Diffusion guidance for inverse problems.}
Diffusion priors can be adapted at inference time by modifying an existing sample, conditioning on measurements, or differentiating through task-specific objectives. SDEdit injects noise into an input and denoises it to balance faithfulness and realism~\citep{meng2022sdedit}; diffusion posterior sampling incorporates gradients of a measurement-consistency objective~\citep{chung2023diffusion}; and training-free guidance generalizes such updates to off-the-shelf differentiable objectives~\citep{ye2024tfg}. Constrained-generation methods likewise use differentiable penalties or {surrogate models} during sampling~\citep{liang2025chance,zampini2026training}. These methods are attractive when the guidance model is reliable along the sampling trajectory. SRD addresses the setting in which the verifier is an expensive, non-differentiable simulator and the differentiable surrogate is useful mainly for proposing where to search.
\\[4pt]
\noindent{\bf Inference-time search and differentiable methods.}
Best-of-$n$ sampling improves generative outputs by drawing several independent
candidates and selecting among them with a scoring model~\citep{ma2025scaling}.
This form of selection is complementary to SRD as best-of-$n$ chooses
among whole trajectories, whereas SRD refines a single one, and the two can be
composed by refining each sampled candidate or by refining only the
best-scoring one. Tree-structured search instead reuses feedback from previous
diffusion rollouts to bias further exploration~\citep{jain2025diffusion}, but
requires additional sampling and reward evaluations, which can be costly when
rewards come from full-wave simulations. SRD focuses on local geometric
refinement within a single diffusion trajectory, coupling each simulator call to
a specific edit. Another class of methods relies on zeroth-order gradient
estimates of smoothed black-box objectives from function
evaluations~\citep{nesterov2017random,zampini2026training,liang2026simulation}. These are useful
when the verifier is inexpensive, but the total cost grows with the number of
sampled directions, and small budgets yield noisy estimates in high-dimensional
layout spaces. Differentiable perturbed optimizers instead smooth and
differentiate optimization maps using random perturbations~\citep{NEURIPS2020_6bb56208}. SRD does not estimate a
dense simulator gradient or differentiate a perturbed optimizer. It searches a
small, signed set of step sizes along one surrogate-proposed geometric
direction.

\section{Settings and Goals}
\label{sec:problem_setup}

\noindent{\bf PCB geometry and response.}
A printed circuit board (PCB) consists of patterned metal layers separated by dielectric substrates and connected by vias.
At specified frequencies, signal transmission is determined by the metal geometry, substrate properties, and port configuration.
We write $\layout\in\mathcal{X}$ for a metal-and-via layout and
$\context\in\mathcal{C}$ for the fixed design context, including port positions, substrate permittivity, and thickness. Together, $(\layout,\context)$ specify the board to be evaluated.
We say that a layout is admissible if it is representable on the prescribed grid, lies within the board outline, and, when combined with the fixed context, yields a geometry that can be evaluated by a full-wave simulator.
In practice, we represent the metal-and-via layout as a grayscale image on a grid, where each pixel encodes the local metal coverage. 
\\[2pt]
For each layout-context pair $(\layout,\context)$, scattering parameters (S-parameters) characterize the frequency-dependent electromagnetic response of the resulting board \citep{pozar2011microwave}. For a board with $P$ ports, $S_{ij}(f)$ is the complex ratio of the outgoing wave amplitude at port $i$ to the incident wave amplitude at port $j$, with all other ports matched. The diagonal terms $S_{ii}(f)$ quantify reflection at each port, whereas the off-diagonal terms $S_{ij}(f)$ quantify transmission between distinct ports. Evaluating every port
pair on a frequency grid $\{f_1,\ldots,f_M\}$ yields a response tensor in $\mathcal{Y}=\mathbb{C}^{P\times P\times M}$.
\\[2pt]
\emph{Inverse design}, the goal of this work, reverses this mapping: given a target response $\target\in\mathcal{Y}$ and a fixed context $\context$, it seeks an admissible layout $\layout$ whose electromagnetic response matches $\target$.

\noindent{\bf Forward models and design losses.}
Computing this response requires solving Maxwell's equations on the board geometry, for example using finite elements or the method of moments \citep{jin2014finite,harrington1993field}. We therefore use a full-wave simulator $\simulator\!:\mathcal{X}\times\mathcal{C}\rightarrow\mathcal{Y}$, as the reference model for evaluating candidate designs. However, each simulation is computationally expensive and thus the community has also developed neural surrogate models $\surrogate\!:\mathcal{X}\times\mathcal{C}\rightarrow\mathcal{Y}$ that provide an inexpensive, differentiable approximation  to the simulator  \citep{dreossi2026inverse,saha2026ktrail}.

A surrogate may be biased and 
% may not preserve the ranking of candidate layouts. 
may not accurately reflect the candidate layout's true S-parameters.
Thus, to evaluate the quality of a layout $\layout$ with respect to a target response $\target$ in context $\context$, we define two loss functions, one based on the simulator and one based on the surrogate.
For a response distance
$\dist:\mathcal{Y}\times\mathcal{Y}\rightarrow\mathbb{R}_{\geq 0}$, define
\begin{equation}
    \simloss(\layout;\target,\context)
    =\dist\!\left(\simulator(\layout,\context),\target\right),
    \qquad
    \surrloss(\layout;\target,\context)
    =\dist\!\left(\surrogate(\layout,\context),\target\right).
    \label{eq:sim_surr_losses}
\end{equation}
The simulator loss measures design quality; the surrogate loss approximates it.

\noindent{\bf Budgeted inverse design.}
Given a target response $\target$, a fixed context $\context$, and a budget $\budget$ of full-wave simulator evaluations, the goal is to find the admissible layout whose simulated response best matches $\target$. 
Let $\numsim(\layout;\target,\context)$ denote the total number of calls to
$\simulator$ used to generate, select, and evaluate a returned layout $\layout$
for the task $(\target,\context)$. We formulate budgeted inverse design as
\begin{equation}
\begin{aligned}
     \layout_{\budget}^{\star}
    \in \arg\min_{\layout\in\mathcal{X}}\quad
    & \simloss(\layout;\target,\context) \\
    \text{subject to}\quad
    & \numsim(\layout;\target,\context)\leq\budget.
\end{aligned}
    \label{eq:objective}
\end{equation}
Here, $\layout_{\budget}^{\star}$ is the best layout attainable under budget $\budget$, and $\simloss$ is the simulator loss defined in Eq.~\ref{eq:sim_surr_losses}. Membership in $\mathcal{X}$ enforces the geometric and simulator-admissibility requirements defined above, whereas the constraint on $\numsim$ limits the expensive full-wave evaluations. %The count $\numsim$ includes calls used for candidate evaluation, selection, and final scoring;
Note that $x$ is a high-dimensional object (an image), and that Eq.~\ref{eq:objective} is a challenging high-dimensional black-box optimization problem, thus motivating the use of generative models to propose candidate layouts.
% Although written as a function of $\layout$, $\numsim$ records the evaluation cost incurred in obtaining that layout, rather than an intrinsic property of its geometry.

\section{Guided Diffusion for Inverse Design}
\label{sec:guided_diffusion}
\label{sec:preliminaries}

\noindent{\bf Conditional diffusion.}
Diffusion models learn a layout prior by reversing a gradual corruption process \citep{ho2020denoising,song2021scorebased}. Let $z_0$ be a continuous representation of a clean layout. Under the standard variance schedule $\alpha_t=1-\beta_t$ and $\bar\alpha_t=\prod_{s\!=\!1}^{t}\alpha_s$, a noisy state is $z_t\!=\!\sqrt{\bar\alpha_t}\,z_0+\sqrt{1-\bar\alpha_t}\,\epsilon$, with $\epsilon\sim\mathcal{N}(0,\mathbf{I})$~\citep{liang2025simultaneous}. A noise predictor $\epsilon_\theta(z_t,t;\target,\context)$, which equivalently parameterizes the conditional score, guides the reverse process from Gaussian noise toward a layout conditioned on the target response and board context. At each step, it yields a clean-layout prediction
\begin{equation}
    z_0^t
    =\frac{z_t-\sqrt{1-\bar\alpha_t}\,
    \epsilon_\theta(z_t,t;\target,\context)}{\sqrt{\bar\alpha_t}}.
    \label{eq:clean_prediction}
\end{equation}
This prediction provides the representation on which design objectives and edits can be evaluated during sampling. It remains continuous; conversion to an admissible PCB is required before full-wave evaluation. Conditioning encourages response matching but does not directly capture the simulator objective in Eq.~\ref{eq:objective}.

\noindent{\bf Loss guidance.}
\label{sec:prelim_guidance}
Guided diffusion supplements the learned denoising process with a
target-dependent loss gradient \citep{chung2023diffusion,ye2024tfg,dreossi2026inverse}.
Because the full-wave simulator does not expose this gradient, a standard
approach uses the differentiable surrogate. Let
$\surrlosscontinuous$ denote its loss extended to the
continuous layout representation, with a differentiable decoder when needed.
The guidance direction is
\begin{equation}
    g_t=-\lambda_t\,\nabla_{z_t}
    \surrlosscontinuous(z_0^t;\target,\context),
    \qquad \lambda_t\geq 0,
    \label{eq:surrogate_guidance}
\end{equation}
where the gradient is taken through the clean prediction and $\lambda_t$
controls the guidance strength. This makes response-directed updates
inexpensive, but requires a differentiable path to the surrogate loss;
hard geometry extraction and binary rasterization do not provide that path.

\noindent{\bf Limitations.}
While differentiability makes guidance computable, it does not ensure that it improves the simulated response. Low surrogate loss can coexist with substantially larger simulator loss (Figure~\ref{fig:overview}(c)). Differentiability alone establishes neither accurate local directions nor reliable step sizes. Optimizing $\surrloss$ can therefore favor layouts that do not improve $\simloss$, particularly when edits leave the surrogate's training distribution.
This motivates using simulator feedback to assess proposed changes while limiting its cost under the budget $\budget$. 

\section{Simulator-Refined Diffusion}
\label{sec:method}

This section introduces \textit{Simulator-Refined Diffusion} (SRD) which leverages a full-wave simulator within the diffusion sampling process and spends its calls on selecting among a small set of candidate edits rather than relying solely on random perturbations. At selected denoising steps, SRD first extracts a geometric representation of the predicted layout and uses the surrogate to propose an edit direction. It then uses simulator feedback from a trial displacement along the surrogate descent direction to determine the subsequent search strategy. If following this direction can sufficiently reduce the simulator loss $\simloss$, SRD tests additional step sizes along the same direction. Otherwise, it explores other directions. The candidate with the lowest simulator loss is then written back into the diffusion state.
The method consists of two key components:
\begin{enumerate}[leftmargin=*,nosep,topsep=0pt]
    \item \textbf{Edit space:} Unlike guidance applied to continuous diffusion states, SRD operates on a low-dimensional geometric representation composed of rectangles. This representation is also required by the simulator and fabrication workflows.
    \item \textbf{Surrogate+simulator edits:} SRD uses surrogate gradients to propose a search direction. The simulator evaluates candidate edits and selects the one with the lowest simulator loss.
    % \item \textbf{Noise-aware scheduling:} Finally, a guidance schedule determines where in the denoising trajectory simulator calls are worth spending.
\end{enumerate}
A key feature of SRD is that it requires no additional training, since the diffusion prior and the surrogate are used as given and guidance happens exclusively at sampling time. This also allows the model to generalize to unseen targets and contexts, since the surrogate is used only to propose edits and the simulator is used to select among them.

Concretely, during generation, at each reverse step $t$ the model predicts a clean sample $z_0^t$ from the current (noisy) state $z_t$. Then, at pre-defined refinement steps, SRD maps this prediction to a structured geometry $p_0 = G(z_0^t)$, which extracts a layout description required by the PCB simulator, queries the surrogate for a direction $u$, and evaluates a small set of candidates induced by $u$ with the simulator. From this set, SRD selects the best candidate $p^\star$, writes it back into the clean prediction, and re-noises
before continuing to $z_{t-1}$:
\begin{tcolorbox}[
  enhanced,
  boxrule=0.6pt,
  colback=gray!8,
  colframe=gray!60,
  arc=1mm,
  boxsep=0.5pt,
  left=0pt,
  right=0pt,
  top=2pt,
  bottom=2pt 
]
\begin{equation*}    
    z_t \;\longrightarrow\; z_0^t \;\xrightarrow{\;G\;} p_0
    \;\xrightarrow{\;\surrogate\;} u
    \;\xrightarrow{\;\simulator\;} p^\star
    \;\longrightarrow\; z_0^t \;\longrightarrow\; z_{t-1}.
\end{equation*}
\end{tcolorbox}
Figure~\ref{fig:overview}, in the Introduction, illustrates schematically this procedure. The following subsections describe the components of SRD in detail. 

\subsection{Structured edit space}
\label{sec:structured_edits}

At each refinement step $t$, the diffusion model predicts a clean layout $z_0^t \in \mathbb{R}^{H\times W}$ from the noisy state $z_t$ using Eq.~\ref{eq:clean_prediction}, where $H$ and $W$ denote the image height and width. The predicted clean sample is a continuous-valued representation of the layout, but the simulator requires an explicit geometric description. To bridge this gap, we define a geometry map $G$ that converts the prediction into a structured representation $p_0=G(z_0^t)$. We implement $G$ using a CenterNet model~\citep{zhou2019objects} trained on rasterized layouts to predict rectangle centers and dimensions.
%The map $G$ is implemented as a differentiable rasterization operator which extracts a description of the metal geometry from the clean prediction and rasterizes it onto the binary grid of admissible layouts: $\layout_t = G(z_0^t)$.

Edits could in principle be applied to $z_0^t$ and mapped through $G$ for evaluation, but continuous state edits result in inefficient allocation of downstream simulator calls. Pixel-space perturbations affect the simulator input through the learned map $G$, so their magnitude does not directly control the resulting changes in rectangle positions or dimensions.
Instead, SRD operates on the structured description that $G$ produces. For a layout with $n$ extracted rectangles, $p_0$ contains the two center coordinates, width, and height of each rectangle. An edit is thus a displacement in $\mathbb{R}^{4n}$ rather than in $\mathbb{R}^{H\times W}$. This representation allows SRD to control rectangle positions and dimensions directly, producing explicit candidate geometries for simulator evaluation.

\subsection{Surrogate-proposed, simulator-selected edits}
\label{sec:propose_select}

Operating in rectangle space significantly reduces the dimensionality and provides direct control over rectangle positions. However, choosing which edits to evaluate remains challenging. Random perturbations can be used to estimate an update direction, but the perturbations themselves are sampled without reference to the design objective. This can squander limited simulator calls on candidates that offer little or no improvement in response matching. To address this issue, SRD uses \textit{the surrogate} to propose an edit direction. In practice, $\surrogate$ is a state-of-the-art forward model pretrained on the same corpus as the diffusion prior (Section~\ref{sec:experiments});
% Let $p_0 = G(z_0^t)$ denote the rectangle parameters extracted from the current clean prediction. 
this surrogate provides a descent direction on the design objective,
\begin{equation}
    u = -\nabla_{p}\,\surrloss(p;\target,\context)\big|_{p=p_0}.
    \label{eq:surrogate_direction}
\end{equation}
% \reviewcomment{Nando}{should we specify where we get this surrogate? LEt's mention that in practice we use a SOTA surrogate model trained on the same data as the diffusion model?}
Computing this direction through the surrogate remains inexpensive in execution since it is natively differentiable, so the cost is negligible in comparison to the full-wave simulator. 
\\[2pt]
Although the surrogate provides a \textit{suggested} direction for improvement, a decrease in surrogate loss may not translate into a decrease in simulator loss (Figure \ref{fig:overview} (c)). Errors in the predicted response can distort the loss gradient, particularly when edits move the geometry away from the training distribution.
To mitigate this issue, SRD first evaluates a trial displacement $p_0 + hu$ with the simulator, where $h$ is a fixed step size. The probe provides direct evidence about this proposed step and determines how subsequent candidates are generated. We consider three ways of using this evidence, which differ in how the remaining candidates are constructed once the probe result is known.
\\[4pt]
The \textbf{directional} variant evaluates additional step sizes along $u$ if the probe sufficiently reduces the simulator loss, and along $-u$ otherwise. The intuition is that the surrogate may identify a useful search axis even when its preferred sign is inaccurate. Reversing the direction allows the simulator to test this possibility while keeping the search focused. However, this strategy cannot explore edits outside the surrogate-defined axis, and a failed probe does not guarantee that the opposite direction will improve the loss.
\\[4pt]
The \textbf{adaptive} variant continues along $u$ only if the probe sufficiently reduces the simulator loss. Otherwise, it falls back to random perturbations of the rectangle parameters. This reduces reliance on the surrogate when its proposed step is ineffective and explores directions outside the initial search axis, at the cost of less focused exploration.
\\[4pt]
Finally, the \textbf{hybrid} variant evaluates both probe-selected directional candidates and random perturbations at each refinement step. Combining these proposals retains the information supplied by the surrogate while considering more diverse geometric edits. This broader search significantly increases the required simulator call budget, but can provide the strongest results in many cases.

For each variant, SRD selects the candidate with the lowest simulator loss from its candidate set $\mathcal{P}$:
\begin{equation}
    p^\star \in \arg\min_{p\in\mathcal{P}}
    \simloss(p;\target,\context).
    \label{eq:candidate_selection}
\end{equation}
Each candidate set includes the unedited geometry $p_0$, ensuring that the selected geometry has a simulator loss no greater than that of $p_0$. Thus, the refinement quality monotonically improves (or remains the same). Algorithm~\ref{alg:srd} (\ref{app:algorithm}) summarizes the full sampling procedure for all three variants.

\textbf{Remark.}
While the search strategies specify how to propose candidate edits, we must decide when to apply refinement and how many candidates to evaluate at each step. These choices are highly coupled under a limited simulator budget.
Earlier refinement can offer greater flexibility, as more denoising steps remain to accommodate layout changes. However, the clean prediction may still be unstable, making rectangle extraction less reliable. Subsequent denoising may also overwrite the selected edit, leading to reduced downstream improvement under a matched budget.
Later refinement operates on a more stable geometry but can reduce opportunities to recover from a poor design.
The candidate set size also introduces a trade-off: larger sets allow more step sizes 
or directions to be compared but require more simulator calls, whereas smaller sets reduce budget cost but may miss useful edits.
In our experiments, SRD consistently outperforms all baselines across the tested refinement settings and candidate budgets (Section~\ref{sec:exp_headline}).

\section{Experiments}
\label{sec:experiments}

\begin{wrapfigure}{r}{0.375\textwidth}
    \centering
    \vspace{-10pt}
    \includegraphics[width=\linewidth]{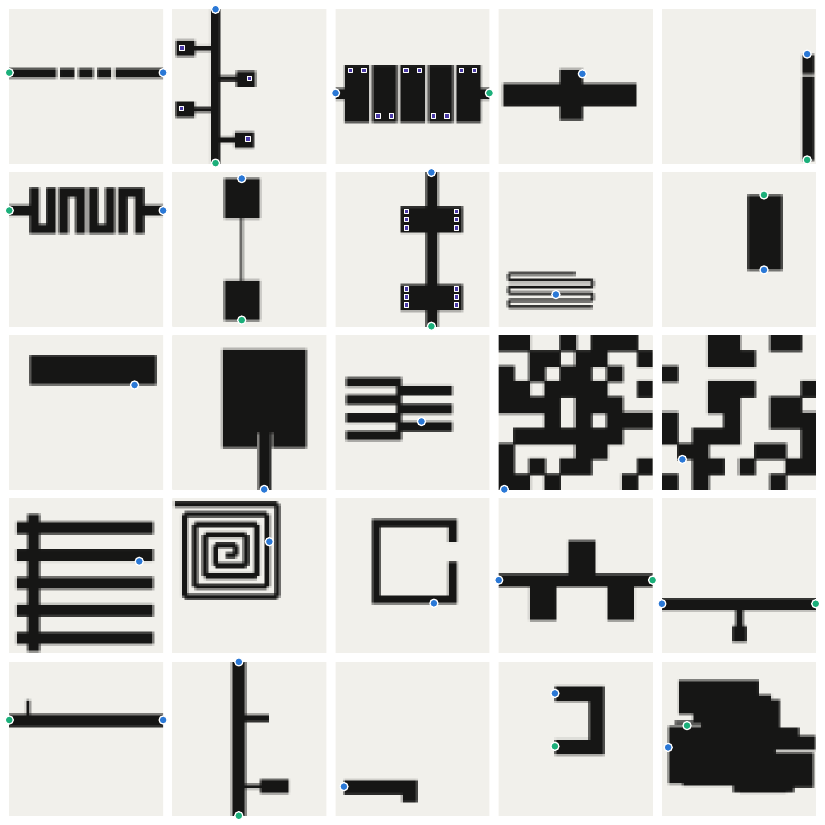}
    \vspace{-8pt}
    \caption{Representative boards.}
    \label{fig:templates_grid_5x5}
\end{wrapfigure}
\textbf{Benchmark.} All methods are evaluated on a test set of 250 samples from 25 layout templates, with 10 samples per template, as shown in Figure~\ref{fig:templates_grid_5x5}. An out-of-distribution (OOD) set of 50 constructed S-parameters is also introduced to expand the range of S-parameters specifications beyond the test set. Each sample specifies a board's complex S-parameters at 51 frequency points over 1--20\,GHz, feed positions, and substrate properties. Both sets include one- and two-port configurations. Ground-truth S-parameters are computed with Palace~\citep{palace}, a full-wave 3D finite-element electromagnetic solver built on the MFEM library~\citep{anderson2021mfem}. Simulations are performed at 51 frequency points over 1--20\,GHz, with a 900\,s time limit per solve. We evaluate generated layouts using the RMSE $\simloss(x;y,c)=\|\simulator(x,c)-y\|_{\mathrm{F}}/(P\sqrt{2M})$, where $M=51$ is the number of frequency points and $P$ is the number of ports. The Frobenius norm is taken over all port pairs and frequencies. Lower values indicate closer agreement with the target.

\textbf{Pretrained models and data.}
All methods share the same two frozen networks from \citet{dreossi2026inverse}.
The diffusion prior is a conditional UNet-based
denoiser over $64{\times}64$ grayscale metal masks of an $8{\times}8$\,mm board,
conditioned on the complex target response (all port pairs at $M{=}51$ frequency
points over 1--20\,GHz), the port positions, and the substrate properties
(permittivity, loss tangent, and thickness). The differentiable surrogate
$\surrogate$ is a ViT-based forward model that maps a layout, ports, and
substrate description to the complex response on the same frequency grid.
Both the surrogate and diffusion models are trained on the same corpus of
$\sim$1.26M procedurally generated PCBs spanning the same family of
parameterized templates, each labeled with full-wave S-parameters obtained from Palace simulations.
Neither network is fine-tuned for any of the following experiments and all guidance
happens at sampling time. Architectures, training hyperparameters, and compute are
detailed in \citet{dreossi2026inverse}.

\textbf{Competing methods.}
We evaluate our method, SRD, alongside three baselines:
\begin{enumerate}[label=(\arabic*), leftmargin=*, nosep,topsep=0pt]
\item \textbf{Conditional diffusion (CD)}~\citep{dreossi2026inverse}: CD generates PCB layouts conditioned on target S-parameters.
\item \textbf{Gradient-guided diffusion (GGD)}~\citep{ye2024tfg}: GGD adapts diffusion posterior sampling to guide denoising with gradients of the S-parameter loss calculated by surrogate.
\item \textbf{Differentiable Perturbed Optimizer (DPO)}~\citep{zampini2026training}: DPO uses random perturbations to estimate update directions from simulator evaluations.
\end{enumerate}
\vspace{-4pt}
Our method, \textbf{Simulator-Refined Diffusion (SRD)}, uses surrogate gradients to propose search directions and simulator evaluations to estimate update directions. The \emph{adaptive} strategy uses perturbations along the surrogate descent direction if an initial probe sufficiently reduces $\simloss$, and random perturbations otherwise. The \emph{directional} strategy follows or reverses the surrogate descent direction based on the probe. The \emph{hybrid} strategy uses these directional perturbations together with random perturbations within each guidance step.

\subsection{Comparison across Methods}
\label{sec:exp_comparison}
\begin{table}[t]
\centering
\fontsize{8}{10}\selectfont
\renewcommand{\arraystretch}{1.12}
\setlength{\tabcolsep}{1.5pt}
\begin{tabularx}{\textwidth}{@{}>{\raggedright\arraybackslash}p{2.8cm}*{2}{>{\centering\arraybackslash}Xcccc}@{}}
\toprule
& \multicolumn{5}{c}{Test set}
& \multicolumn{5}{c}{OOD set} \\
\cmidrule(lr){2-6}
\cmidrule(l){7-11}
\multicolumn{1}{@{}c}{Method}
& \shortstack[c]{Mean\\loss$\downarrow$}
& \shortstack[c]{Median\\loss$\downarrow$}
& \shortstack[c]{Reduction\\(\%)$\uparrow$}
& \shortstack[c]{Sim. success\\(\%)$\uparrow$}
& \shortstack[c]{Palace\\calls}
& \shortstack[c]{Mean\\loss$\downarrow$}
& \shortstack[c]{Median\\loss$\downarrow$}
& \shortstack[c]{Reduction\\(\%)$\uparrow$}
& \shortstack[c]{Sim. success\\(\%)$\uparrow$}
& \shortstack[c]{Palace\\calls} \\
\midrule
CD~{\scriptsize\citep{dreossi2026inverse}}
& 0.3222 & 0.2820 & 0.00 & \underline{99.6} & 1
& 0.5418 & 0.6120 & 0.00 & \textbf{98.0} & 1 \\
GGD~{\scriptsize\citep{ye2024tfg}}
& 0.3174 & 0.2687 & 1.49 & 98.4 & 1
& 0.5249 & 0.6121 & 3.12 & \underline{96.0} & 1 \\
DPO~{\scriptsize\citep{zampini2026training}}
& 0.2819 & 0.2441 & 12.51 & \textbf{100.0} & 7
& 0.4821 & 0.5403 & 11.02 & \underline{96.0} & 17 \\
\midrule
\rowcolor[gray]{0.90}
\textbf{SRD (adaptive)}
& \underline{0.2608} & \textbf{0.1977} & \underline{19.06} & \textbf{100.0} & 7
& 0.4444 & \underline{0.4338} & 17.98 & \textbf{98.0} & 17 \\
\rowcolor[gray]{0.90}
\mbox{\textbf{SRD (directional)}}
& 0.2640 & \underline{0.1980} & 18.06 & \textbf{100.0} & 7
& \textbf{0.4346} & \textbf{0.4082} & \textbf{19.79} & \textbf{98.0} & 17 \\
\rowcolor[gray]{0.90}
\textbf{SRD (hybrid)}
& \textbf{0.2539} & \textbf{0.1977} & \textbf{21.20} & \textbf{100.0} & 11
& \underline{0.4361} & 0.4368 & \underline{19.51} & \underline{96.0} & 29 \\
\bottomrule
\end{tabularx}
\vspace{-4pt}
\caption{Comparison on the test and OOD sets. DPO and all SRD variants apply guidance during the last 2 denoising steps on the test set and the last 4 on the OOD set.
Palace calls count the selection budget at each guided step plus one final Palace evaluation; CD and GGD use only the final evaluation.
Reductions are relative to the mean loss of CD. Simulation success is the percentage of generated boards with a valid Palace result. For each performance metric, bold and underlined values indicate the best and second-best results, respectively.}
\label{tab:method_comparison}
\end{table}
Table~\ref{tab:method_comparison} shows that all SRD variants outperform the baselines in both mean and median S-parameter loss $\simloss$ on both sets. GGD provides only modest loss reductions over CD (1.49\% and 3.12\%), whereas DPO achieves larger improvements (12.51\% and 11.02\%) through simulator feedback. With the same Palace call budgets as DPO, SRD increases these reductions to 19.06\% and 17.98\% on the test and OOD sets, respectively. These results demonstrate that surrogate gradients help identify more promising candidate edits than random perturbations, while simulator evaluations determine which edits most effectively reduce the S-parameter loss.

\Needspace{42\baselineskip}
\begin{wrapfigure}[35]{r}{0.525\textwidth}
    \centering
    \vspace{-2pt}
    \includegraphics[width=\linewidth]{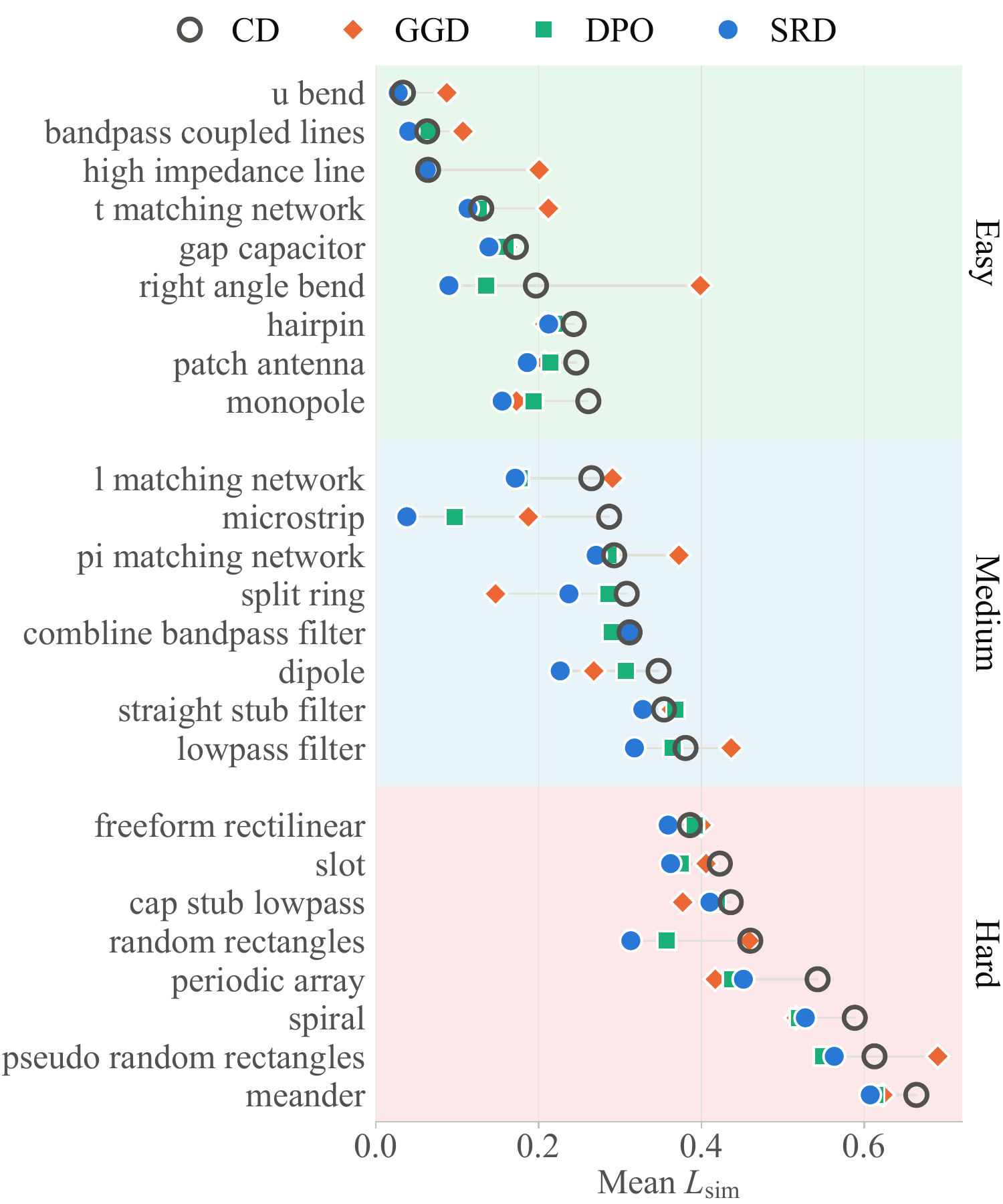}
    \setlength{\abovecaptionskip}{4pt}
    \vspace{-12pt}
    \caption{Mean S-parameter loss by template on the test set. Shaded regions indicate the easy, medium, and hard target groups. Lower loss is better.}
    \par \vspace{4pt}
    \label{fig:per_template_lsim}
    \includegraphics[width=\linewidth]{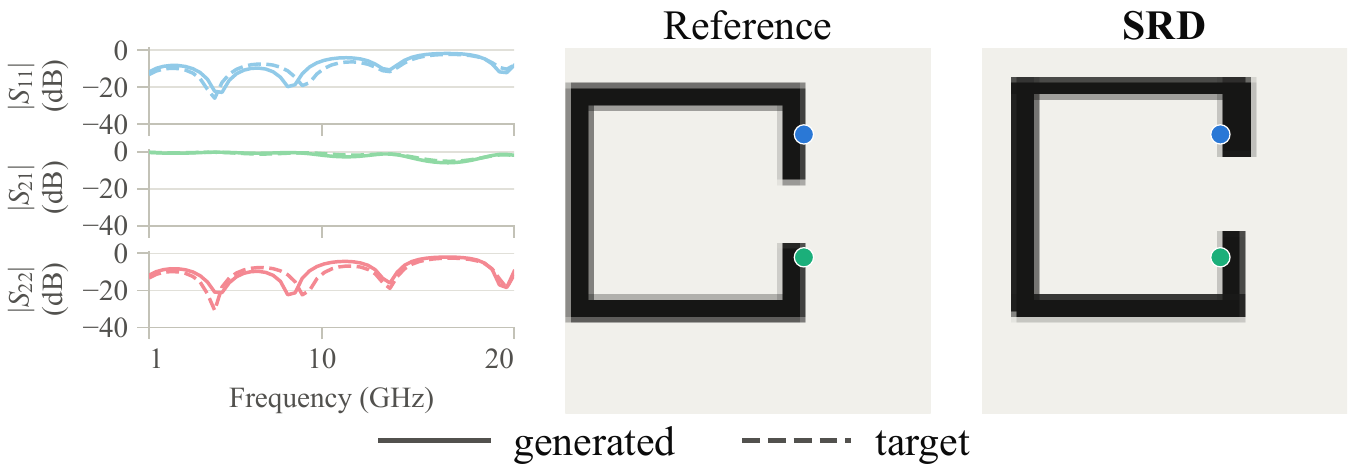}
    \vspace{-12pt}
    \caption{Reference (center) and SRD (right) layouts with their target S-parameters (left).}
    \label{fig:t1704_comparison}
    \vspace{-12pt}
\end{wrapfigure}
On the test set, SRD (hybrid) considers both perturbations along the surrogate direction and random perturbations within each guidance step. This larger candidate set helps identify better edits, yielding the lowest mean S-parameter loss. On the OOD set, the three variants perform similarly, with SRD (directional) achieving the lowest mean loss. The diffusion model cannot produce favorable samples with low S-parameter loss before guidance is applied. Therefore, larger perturbations may offer greater opportunities for improvement. SRD spends part of its Palace call budget assessing whether to follow the surrogate direction, leaving fewer calls to evaluate larger perturbations. This allocation induces slightly higher loss than the directional and hybrid variants.

Next, Figure~\ref{fig:per_template_lsim} compares the mean S-parameter loss across 25 templates grouped by difficulty. SRD's gains over CD are most pronounced on medium and hard templates, where CD produces larger response errors. SRD also improves on easy templates, although the loss levels reached by CD are already low, leaving relatively less room for improvement.

Finally, we report an example layout generated by SRD alongside the reference layout and their S-parameter responses target (Figure~\ref{fig:t1704_comparison}). Notice how the SRD proposed layout produces responses that are closer to the target values (Figure~\ref{fig:t1704_comparison}(left)).
Interestingly, when comparing the reference and the generated layouts, one can note that while the port locations remain fixed, the metal geometry generated shifts relative to them. In the reference layout, the ports lie on the right edges of the vertical metal segments. In the SRD layout, they lie on the left edges. Throughout the generated set, we observe similar results. This artifact illustrates the ``creative'' ability of SRD in searching for good layouts that achieve the target S-parameter response. In other words, SRD is capable of constructing novel layouts for a given design objective.
\\[2pt]
Appendix~\ref{app:perturbations} illustrates how SRD perturbs the layout during the denoising steps. Appendix~\ref{app:prior_proximity} further examines how SRD improves response accuracy while retaining proximity to the learning distribution. Appendix~\ref{app:qualitative_comparison} provides further qualitative comparisons across methods and SRD.

\subsection{Quality--Cost Trade-off}
\label{sec:exp_headline}

\begin{wrapfigure}[11]{r}{0.65\textwidth}
    \centering
    \vspace{-14pt}
    \includegraphics[width=\linewidth]{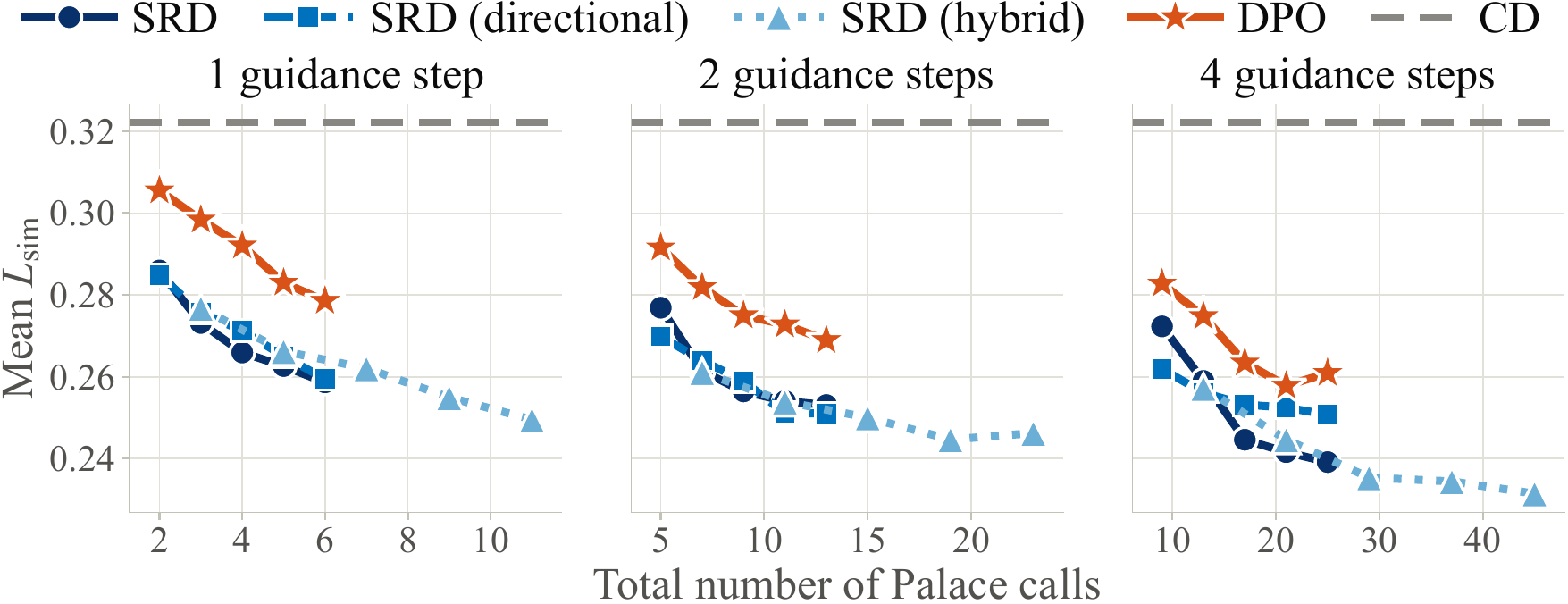}
    \vspace{-18pt}
    \caption{Quality--cost trade-off across methods.}
    \label{fig:budget_grid}

\end{wrapfigure}
Figure~\ref{fig:budget_grid} shows that all SRD variants consistently outperform CD and DPO in three different number of guidance steps settings. SRD (hybrid) achieves the lowest mean S-parameter loss in each setting with a larger Palace call budget, which allows evaluating both directional and random perturbations.
The right panel illustrates the efficiency of adaptive strategy selection. At 21 Palace calls, SRD achieves lower loss than both variants by adaptively selecting the more effective search strategy. However, it performs slightly worse than either at 13 calls because each guidance step uses two initial calls to evaluate the current design and a probe, leaving fewer calls for further exploration.
Furthermore, the middle and right panels show how budget allocation affects performance. At nine Palace calls, SRD achieves lower loss with two guidance steps than with four. Under this tight budget, SRD can benefit more from evaluating additional candidates per step than from increasing the number of guidance steps.

\subsection{Ablations}
\label{sec:exp_ablations}
To evaluate the contributions of surrogate directions and simulator feedback, we vary the perturbation strategy and candidate selector in Table~\ref{tab:ablations}. With random perturbations, surrogate-based selection yields a mean loss of 0.3263, slightly above CD (0.3222), while Palace-based selection reduces it to 0.2819. This comparison supports the use of simulator feedback for selecting effective edits, since surrogate predictions may not reliably track the simulated loss.

Surrogate gradients also improve candidate generation. Under surrogate-based selection, surrogate descent directions yield a mean loss of 0.2873, compared with 0.3263 for random perturbations. SRD (adaptive) combines surrogate-proposed directions with Palace-based selection and a conditional fallback to random perturbations. It achieves the lowest mean and median losses among the configurations in this ablation: 0.2608 and 0.1977, respectively.

\begin{table}[t]
\centering
\begin{minipage}{0.94\linewidth}
\small
\fontsize{8.5}{10}\selectfont
\renewcommand{\arraystretch}{1.18}
\setlength{\tabcolsep}{4pt}
\setlength{\heavyrulewidth}{0.65pt}
\newcommand{\ablationconditionalcheck}{%
    \tikz[baseline=(mark.base)]{%
        \node[inner sep=0pt] (mark) {$\checkmark$};
        \node[inner sep=0pt] at (mark.center) {$\times$};}}
\begin{tabularx}{\linewidth}{>{\raggedright\arraybackslash}p{1.45cm}*{4}{>{\centering\arraybackslash}X}rrr}
\toprule
& \multicolumn{2}{c}{\textbf{Perturbations}}
& \multicolumn{2}{c}{\textbf{Candidate selection}}
& \multicolumn{3}{c}{\textbf{Performance}} \\
\cmidrule(lr){2-3}
\cmidrule(lr){4-5}
\cmidrule(lr){6-8}
Method
& \shortstack[c]{Surrogate\\direction}
& Random
& Palace
& Surrogate
& \shortstack[r]{Mean\\loss\,$\downarrow$}
& \shortstack[r]{Median\\loss\,$\downarrow$}
& \shortstack[r]{Reduction\\(\%)\,$\uparrow$} \\
\midrule
\multirow{3}{*}{Ablations}
& $\times$ & $\checkmark\!\!\checkmark$ & $\times$ & $\times$
& 0.4064 & 0.4058 & $-26.14$ \\
& $\times$ & $\checkmark\!\!\checkmark$ & $\times$ & $\checkmark\!\!\checkmark$
& 0.3263 & 0.2986 & $-1.28$ \\
& $\checkmark\!\!\checkmark$ & $\times$ & $\times$ & $\checkmark\!\!\checkmark$
& 0.2873 & 0.2284 & 10.83 \\
\midrule
CD
& $\times$ & $\times$ & $\times$ & $\times$
& 0.3222 & 0.2820 & 0.00 \\
DPO
& $\times$ & $\checkmark\!\!\checkmark$ & $\checkmark\!\!\checkmark$ & $\times$
& 0.2819 & 0.2441 & 12.51 \\
\rowcolor[gray]{0.93}
\textbf{SRD}
& $\checkmark\!\!\checkmark$ & $\checkmark$ & $\checkmark\!\!\checkmark$ & $\times$
& \textbf{0.2608} & \textbf{0.1977} & \textbf{19.06} \\
\bottomrule
\end{tabularx}
\end{minipage}
\caption{Ablations on the test set. Double checkmarks indicate components that are always used. A single checkmark indicates conditional use, and crosses indicate inactive components. Candidates use surrogate descent directions or random perturbations; selection uses Palace evaluations or surrogate predictions. Mean loss reductions are relative to CD.}
\label{tab:ablations}
\end{table}

\section{Limitations and Conclusion}
\label{sec:conclusion}
This paper introduced Simulator-Refined Diffusion (SRD) for radio-frequency printed circuit board (PCB) inverse design. 
SRD is a diffusion model that incorporates surrogate-based search directions and full-wave simulator feedback. At selected denoising steps, SRD extracts a structured geometric representation of the predicted clean layout and uses a surrogate gradient to propose an edit direction. Then, a full-wave simulator probe assesses this direction and guides the generation of candidate edits. %SRD then writes the candidate with the lowest simulator loss back into the diffusion state before resuming denoising.
Across 250 test targets spanning 25 layout templates and 50 constructed out-of-distribution targets, SRD variants reduce mean simulated S-parameter loss by up to 21.2\% and 19.8\%, respectively, relative to a state-of-the-art conditional diffusion model for PCB design. Our proposed approach also outperforms other baselines at matched simulator-call budgets on both sets and an extensive ablation study supported the complementary roles of surrogate directions and simulator-based selection. These results are significant and suggest that the dual use of the surrogate and simulator feedback can become an important tool for the PCB inverse design process.
% suggest that a surrogate can provide informative search directions, enabling more accurate response matching in diffusion-based RF design.
\\[2pt]
Building on these promising results, future work could extend SRD's rectangle-based representation to capture finer geometric features and more complex topologies. Additionally, exploring search directions beyond those proposed by the surrogate could further expand the range of PCB designs the method can produce.

\clearpage
% Required and recommended statements (do not count toward the page limit).

\subsection*{AI use statement}
% REQUIRED by ICLR 2027; max 1 page. See the ICLR 2027 AI Policy for Authors.
% TODO fill in: tasks with required disclosure, tasks not used for, recommended
% disclosures, and how AI-assisted work was reviewed/verified.
% \reviewdel{Placeholder.}

In this paper, we used AI tools to assist in creating assets for diagrams and figures, for accelerating code development pipelines, and in a limited role for refining the written exposition. We have not used generative AI tools in any way to support development of the methodology, in any role contributing to mathematical correctness of the work, nor as an author of the paper. All AI assisted work has been carefully reviewed, and we take full responsibility for the content of the work.

\subsection*{Ethics statement}
% Recommended; max 1 page.

This work studies computational inverse design of RF printed-circuit-board layouts and does not introduce a dataset of people or make decisions about individuals. Its intended benefit is to improve electromagnetic-response matching under a constrained full-wave simulation budget. Potential risks include dual-use applications of improved RF design tools and the computational and energy cost of model training and repeated finite-element simulation. The method also remains subject to simulation error and should not be treated as sufficient validation for fabrication or safety-critical deployment. Appropriate downstream use requires domain review, manufacturing checks, regulatory compliance where applicable, and physical validation.

\subsection*{Reproducibility statement}
% Recommended. Reference (don't restate) the parts of the paper/appendix/supplement
% that support reproducibility: code link, simulator settings, dataset processing.
The problem definition and budget formulation are given in Section~\ref{sec:problem_setup}; the sampling procedure is specified in Section~\ref{sec:method} and Algorithm~\ref{alg:srd}; and the benchmark, Palace metric, baselines, and guidance settings are described in Section~\ref{sec:experiments} and Table~\ref{tab:method_comparison}. Appendix~\ref{app:simulator} records the available simulator settings, and Appendix~\ref{app:algorithm} gives the complete sampling pseudocode.

\bibliography{iclr2027_conference}

\begin{thebibliography}{27}
\providecommand{\natexlab}[1]{#1}
\providecommand{\url}[1]{\texttt{#1}}
\expandafter\ifx\csname urlstyle\endcsname\relax
  \providecommand{\doi}[1]{doi: #1}\else
  \providecommand{\doi}{doi: \begingroup \urlstyle{rm}\Url}\fi

\bibitem[Aage \& Johansen(2017)Aage and Johansen]{aage2017topology}
Niels Aage and Villads~Egede Johansen.
\newblock Topology optimization of microwave waveguide filters.
\newblock \emph{International Journal for Numerical Methods in Engineering},
  112\penalty0 (3):\penalty0 283--300, 2017.
\newblock \doi{10.1002/nme.5551}.
\newblock URL \url{https://doi.org/10.1002/nme.5551}.

\bibitem[Anderson et~al.(2021)Anderson, Andrej, Barker, Bramwell, Camier,
  Cerveny, Dobrev, Dudouit, Fisher, Kolev, et~al.]{anderson2021mfem}
Robert Anderson, Julian Andrej, Andrew Barker, Jamie Bramwell, Jean-Sylvain
  Camier, Jakub Cerveny, Veselin Dobrev, Yohann Dudouit, Aaron Fisher, Tzanio
  Kolev, et~al.
\newblock Mfem: A modular finite element methods library.
\newblock \emph{Computers \& Mathematics with Applications}, 81:\penalty0
  42--74, 2021.

\bibitem[{AWS Center for Quantum Computing}(2023)]{palace}
{AWS Center for Quantum Computing}.
\newblock {Palace}: {PArallel} {LArge}-scale {Computational}
  {Electromagnetics}, 2023.
\newblock URL \url{https://github.com/awslabs/palace}.

\bibitem[Berthet et~al.(2020)Berthet, Blondel, Teboul, Cuturi, Vert, and
  Bach]{NEURIPS2020_6bb56208}
Quentin Berthet, Mathieu Blondel, Olivier Teboul, Marco Cuturi, Jean-Philippe
  Vert, and Francis Bach.
\newblock Learning with differentiable perturbed optimizers.
\newblock In H.~Larochelle, M.~Ranzato, R.~Hadsell, M.F. Balcan, and H.~Lin
  (eds.), \emph{Advances in Neural Information Processing Systems}, volume~33,
  pp.\  9508--9519. Curran Associates, Inc., 2020.
\newblock URL
  \url{https://proceedings.neurips.cc/paper_files/paper/2020/file/6bb56208f672af0dd65451f869fedfd9-Paper.pdf}.

\bibitem[Chung et~al.(2023)Chung, Kim, Mccann, Klasky, and
  Ye]{chung2023diffusion}
Hyungjin Chung, Jeongsol Kim, Michael~Thompson Mccann, Marc~Louis Klasky, and
  Jong~Chul Ye.
\newblock Diffusion posterior sampling for general noisy inverse problems.
\newblock In \emph{The Eleventh International Conference on Learning
  Representations}, 2023.
\newblock URL \url{https://openreview.net/forum?id=OnD9zGAGT0k}.

\bibitem[Dreossi et~al.(2026)Dreossi, Bryant, Liu, Mirman, Kessler, Frei, and
  Krishnaswamy]{dreossi2026inverse}
Tommaso Dreossi, Christopher~M Bryant, Hao Liu, Nathan Mirman, Noah Kessler,
  Michael Frei, and Harish Krishnaswamy.
\newblock Inverse design of multi-layer sub-pixel-resolution rf passives
  through grayscale diffusion with flexible s-parameter conditioning.
\newblock \emph{arXiv preprint arXiv:2605.08233}, 2026.

\bibitem[Geuzaine \& Remacle(2009)Geuzaine and Remacle]{geuzaine2009gmsh}
Christophe Geuzaine and Jean-Fran{\c{c}}ois Remacle.
\newblock Gmsh: A 3-d finite element mesh generator with built-in pre-and
  post-processing facilities.
\newblock \emph{International journal for numerical methods in engineering},
  79\penalty0 (11):\penalty0 1309--1331, 2009.

\bibitem[Guo et~al.(2025)Guo, Karahan, Li, Shao, Zhang, Wang, and
  Sengupta]{11103838}
Yingqing Guo, Emir~Ali Karahan, Zihao Li, Zijian Shao, Zaixi Zhang, Mengdi
  Wang, and Kaushik Sengupta.
\newblock Dall-em: Generative ai with diffusion models for new design space
  discovery and target-to-electromagnetic structure synthesis.
\newblock In \emph{2025 IEEE/MTT-S International Microwave Symposium - IMS
  2025}, pp.\  926--929, 2025.
\newblock \doi{10.1109/IMS40360.2025.11103838}.

\bibitem[Harrington(1993)]{harrington1993field}
Roger~F. Harrington.
\newblock \emph{Field Computation by Moment Methods}.
\newblock Wiley-IEEE Press, 1993.
\newblock ISBN 978-0-7803-1014-8.
\newblock Reissue of the 1968 edition.

\bibitem[Hassan et~al.(2014)Hassan, Wadbro, and Berggren]{hassan2014topology}
Emadeldeen Hassan, Eddie Wadbro, and Martin Berggren.
\newblock Topology optimization of metallic antennas.
\newblock \emph{IEEE Transactions on Antennas and Propagation}, 62\penalty0
  (5):\penalty0 2488--2500, 2014.
\newblock \doi{10.1109/TAP.2014.2309112}.
\newblock URL \url{https://ieeexplore.ieee.org/document/6750741}.

\bibitem[Ho et~al.(2020)Ho, Jain, and Abbeel]{ho2020denoising}
Jonathan Ho, Ajay Jain, and Pieter Abbeel.
\newblock Denoising diffusion probabilistic models.
\newblock In H.~Larochelle, M.~Ranzato, R.~Hadsell, M.~F. Balcan, and H.~Lin
  (eds.), \emph{Advances in Neural Information Processing Systems}, volume~33,
  pp.\  6840--6851. Curran Associates, Inc., 2020.
\newblock URL
  \url{https://proceedings.neurips.cc/paper/2020/hash/4c5bcfec8584af0d967f1ab10179ca4b-Abstract.html}.

\bibitem[Jain et~al.(2025)Jain, Sareen, Pedramfar, and
  Ravanbakhsh]{jain2025diffusion}
Vineet Jain, Kusha Sareen, Mohammad Pedramfar, and Siamak Ravanbakhsh.
\newblock Diffusion tree sampling: Scalable inference-time alignment of
  diffusion models.
\newblock \emph{Advances in Neural Information Processing Systems},
  38:\penalty0 161584--161623, 2025.
\newblock \doi{10.52202/085713-4871}.
\newblock URL
  \url{https://proceedings.neurips.cc/paper_files/paper/2025/hash/d6484394c4cb5e1f4ecad8d90b912025-Abstract-Conference.html}.

\bibitem[Jin(2014)]{jin2014finite}
Jian-Ming Jin.
\newblock \emph{The Finite Element Method in Electromagnetics}.
\newblock Wiley-IEEE Press, 3 edition, 2014.
\newblock ISBN 978-1-118-57136-1.

\bibitem[Liang et~al.(2025{\natexlab{a}})Liang, Christopher, Koenig, and
  Fioretto]{liang2025simultaneous}
Jinhao Liang, Jacob~K Christopher, Sven Koenig, and Ferdinando Fioretto.
\newblock Simultaneous multi-robot motion planning with projected diffusion
  models.
\newblock \emph{arXiv preprint arXiv:2502.03607}, 2025{\natexlab{a}}.

\bibitem[Liang et~al.(2025{\natexlab{b}})Liang, Sun, Samaddar, Madireddy, and
  Fioretto]{liang2025chance}
Jinhao Liang, Yixuan Sun, Anirban Samaddar, Sandeep Madireddy, and Ferdinando
  Fioretto.
\newblock Chance-constrained flow matching for high-fidelity constraint-aware
  generation.
\newblock \emph{arXiv preprint arXiv:2509.25157}, 2025{\natexlab{b}}.

\bibitem[Liang et~al.(2026)Liang, Koenig, and Fioretto]{liang2026simulation}
Jinhao Liang, Sven Koenig, and Ferdinando Fioretto.
\newblock Simulation-informed diffusion for decentralized multi-robot motion
  planning.
\newblock \emph{arXiv preprint arXiv:2605.27697}, 2026.

\bibitem[Ma et~al.(2025)Ma, Tong, Jia, Hu, Su, Zhang, Yang, Li, Jaakkola, Jia,
  and Xie]{ma2025scaling}
Nanye Ma, Shangyuan Tong, Haolin Jia, Hexiang Hu, Yu-Chuan Su, Mingda Zhang,
  Xuan Yang, Yandong Li, Tommi Jaakkola, Xuhui Jia, and Saining Xie.
\newblock Scaling inference time compute for diffusion models.
\newblock In \emph{Proceedings of the IEEE/CVF Conference on Computer Vision
  and Pattern Recognition}, pp.\  2523--2534, 2025.
\newblock URL
  \url{https://openaccess.thecvf.com/content/CVPR2025/html/Ma_Scaling_Inference_Time_Compute_for_Diffusion_Models_CVPR_2025_paper.html}.

\bibitem[Meng et~al.(2022)Meng, He, Song, Song, Wu, Zhu, and
  Ermon]{meng2022sdedit}
Chenlin Meng, Yutong He, Yang Song, Jiaming Song, Jiajun Wu, Jun-Yan Zhu, and
  Stefano Ermon.
\newblock {SDEdit}: Guided image synthesis and editing with stochastic
  differential equations.
\newblock In \emph{International Conference on Learning Representations}, 2022.
\newblock URL \url{https://openreview.net/forum?id=aBsCjcPu_tE}.

\bibitem[Nesterov \& Spokoiny(2017)Nesterov and Spokoiny]{nesterov2017random}
Yurii Nesterov and Vladimir Spokoiny.
\newblock Random gradient-free minimization of convex functions.
\newblock \emph{Foundations of Computational Mathematics}, 17\penalty0
  (2):\penalty0 527--566, 2017.
\newblock \doi{10.1007/s10208-015-9296-2}.
\newblock URL \url{https://doi.org/10.1007/s10208-015-9296-2}.

\bibitem[Pozar(2011)]{pozar2011microwave}
David~M. Pozar.
\newblock \emph{Microwave Engineering}.
\newblock John Wiley \& Sons, 4 edition, 2011.
\newblock ISBN 978-0-470-63155-3.

\bibitem[Requicha(1980)]{requicha1980representations}
Aristides~G Requicha.
\newblock Representations for rigid solids: Theory, methods, and systems.
\newblock \emph{ACM Computing Surveys (CSUR)}, 12\penalty0 (4):\penalty0
  437--464, 1980.

\bibitem[Saha et~al.(2026)Saha, Newell, O'Leary, Natarajan, and
  Aghasi]{saha2026ktrail}
Piyush Saha, Evan Newell, Hanna O'Leary, Arun Natarajan, and Alireza Aghasi.
\newblock {K-TRAIL}: Simulator-guided generative design of {EM/RF} circuits,
  2026.
\newblock URL \url{https://arxiv.org/abs/2609.23183}.

\bibitem[Sim{\'e}oni et~al.(2025)Sim{\'e}oni, Vo, Seitzer, Baldassarre, Oquab,
  Jose, Khalidov, Szafraniec, Yi, Ramamonjisoa, et~al.]{simeoni2025dinov3}
Oriane Sim{\'e}oni, Huy~V Vo, Maximilian Seitzer, Federico Baldassarre, Maxime
  Oquab, Cijo Jose, Vasil Khalidov, Marc Szafraniec, Seungeun Yi, Micha{\"e}l
  Ramamonjisoa, et~al.
\newblock Dinov3.
\newblock \emph{arXiv preprint arXiv:2508.10104}, 2025.

\bibitem[Song et~al.(2021)Song, Sohl-Dickstein, Kingma, Kumar, Ermon, and
  Poole]{song2021scorebased}
Yang Song, Jascha Sohl-Dickstein, Diederik~P. Kingma, Abhishek Kumar, Stefano
  Ermon, and Ben Poole.
\newblock Score-based generative modeling through stochastic differential
  equations.
\newblock In \emph{International Conference on Learning Representations}, 2021.
\newblock URL \url{https://openreview.net/forum?id=PxTIG12RRHS}.

\bibitem[Ye et~al.(2024)Ye, Lin, Han, Xu, Liu, Liang, Ma, Zou, and
  Ermon]{ye2024tfg}
Haotian Ye, Haowei Lin, Jiaqi Han, Minkai Xu, Sheng Liu, Yitao Liang, Jianzhu
  Ma, James Zou, and Stefano Ermon.
\newblock Tfg: Unified training-free guidance for diffusion models.
\newblock \emph{Advances in Neural Information Processing Systems},
  37:\penalty0 22370--22417, 2024.

\bibitem[Zampini et~al.(2025)Zampini, Christopher, Oneto, Anguita, and
  Fioretto]{zampini2026training}
Stefano Zampini, Jacob~K Christopher, Luca Oneto, Davide Anguita, and
  Ferdinando Fioretto.
\newblock Training-free constrained generation with stable diffusion models.
\newblock \emph{Advances in Neural Information Processing Systems},
  38:\penalty0 30993--31024, 2025.
\newblock \doi{10.52202/085713-0918}.
\newblock URL
  \url{https://proceedings.neurips.cc/paper_files/paper/2025/hash/2774a3b52d436b5930da660dd2b32b3a-Abstract-Conference.html}.

\bibitem[Zhou et~al.(2019)Zhou, Wang, and Kr{\"a}henb{\"u}hl]{zhou2019objects}
Xingyi Zhou, Dequan Wang, and Philipp Kr{\"a}henb{\"u}hl.
\newblock Objects as points.
\newblock \emph{arXiv preprint arXiv:1904.07850}, 2019.

\end{thebibliography}
\bibliographystyle{iclr2027_conference}

\appendix
\section{Appendix}
\label{sec:appendix}

\subsection{Simulator configuration and evaluation}
\label{app:simulator}
Candidate and final layouts are converted into solid models~\citep{requicha1980representations}, meshed with \texttt{gmsh}~\citep{geuzaine2009gmsh}, and evaluated with Palace~\citep{palace}. Each Palace invocation returns the complete S-matrix for all ports at 51 uniformly spaced frequencies from 1 to 20\,GHz; internal frequency-sweep solves are included in that invocation.

\paragraph{Geometry and materials.}
Boards measure $8\times8$\,mm and use a $64\times64$ raster. CenterNet~\citep{zhou2019objects} extracts the rectangular signal geometry. The signal layer and full-board ground plane are each 35\,$\mu$m thick and modeled as conductive copper volumes with $\sigma=5.8\times10^7$\,S/m. Substrate relative permittivity $\varepsilon_r$, dielectric loss tangent $\tan\delta$, and thickness are listed in Table~\ref{tab:em-substrates}. Vias have a 0.3\,mm diameter and a 0.6\,mm pad diameter.

\begin{table}[htbp]
\centering
\small
\caption{Substrate properties used in simulation.}
\label{tab:em-substrates}
\begin{tabular}{lccr}
\toprule
Material & $\varepsilon_r$ & $\tan\delta$ & Thickness ($\mu$m) \\
\midrule
Air (suspended) & 1.00 & 0      & 203.2 \\
Rogers RO4003   & 3.55 & 0.0027 & 203.2 \\
FR-4 (S1000H)   & 4.60 & 0.011  & 200.0 \\
\bottomrule
\end{tabular}
\end{table}

\paragraph{Ports and boundaries.}
Each feed uses a 50\,$\Omega$ lumped port spanning the substrate between the ground and signal layers, with width $\min(0.45\,\mathrm{mm},\text{trace width})$ and excitation along $+z$. The air box extends 2\,mm laterally and above the board, with no air below it. The five side and top faces use second-order absorbing conditions; the bottom face has no explicit boundary tag and uses the solver's natural perfect-magnetic-conductor condition.

\paragraph{Mesh and solver.}
We use \texttt{gmsh} 4.15.2 and Palace v0.16.0. Meshes use first-order tetrahedral geometry with characteristic lengths of 50\,$\mu$m--1\,mm and no adaptive mesh refinement. Failed meshing attempts are retried with a 30\,$\mu$m minimum size and then uniform sizing. Palace uses order-2 N\'ed\'elec elements, the SuperLU\_DIST direct solver, and an adaptive reduced-order frequency sweep with tolerance $10^{-2}$. Each solve uses one MPI rank on an NVIDIA A10G GPU with a 900\,s timeout.

% \FloatBarrier

% \reviewadd{\subsection{Simulator-call accounting}
% \label{app:budget_accounting}}

% \reviewadd{The primary budget is the number of Palace invocations per target. The candidate counts in Table~\ref{tab:method_comparison} specify nominal candidates per guidance step rather than total Palace invocations. Parallel candidate evaluations reduce sequential latency only when sufficient concurrent hardware is available; they do not reduce the total-call budget.}

% \reviewcomment{Revision}{Verify the preceding seven-call decomposition against the implementation. State which evaluations of $P_0$, $P_1$, and the candidate set in Algorithm~\ref{alg:simguide_complete} are reused, how duplicate layouts are detected, whether a previously simulated selected candidate is simulated again for final scoring, and how invalid or timed-out candidates consume the budget. Provide total calls, sequential rounds, and wall-clock time for every method and budget.}

% \Needspace{44\baselineskip}

\subsection{Complete sampling algorithm}
\label{app:algorithm}

\begin{algorithm}[H]
\caption{SRD Sampling}
\label{alg:srd}
\label{alg:simguide_complete}
\small
\algrenewcommand\algorithmicindent{1.0em}
\algrenewcommand\alglinenumber[1]{\small #1:}
\begin{algorithmic}[1]
\Require $\target,\context$; $K$ reverse sampling steps $t_K>\cdots>t_1$; refinement at the final $N$ steps; step size $h$
\Statex absolute loss reduction threshold $\varepsilon_{\mathrm{imp}}$; budget $\budget$; variant $v$; candidate counts $m_{\mathrm d},m_{\mathrm r}$
\Ensure PCB layout $\widehat{\layout}$

\State $z\sim\mathcal N(0,\mathbf I)$
\For{$k=K,\ldots,1$}
    \State $\epsilon\gets\epsilon_\theta(z,t_k;\target,\context)$
    \State $z_0^{t_k}\gets\Call{PredictClean}{z,\epsilon,t_k}$

    \If{$k\leq N$ \textbf{and} budget permits refinement}
        \State $p_0\gets G(z_0^{t_k})$
        \State $u\gets-\nabla_p\surrloss
            (p;\target,\context)|_{p=p_0}$
        \State $p_1\gets p_0+hu$
        \State $(\ell_0,\ell_1)\gets\Call{Simulate}{p_0,p_1}$
        \State $s\gets+1$ \textbf{if}
            $\ell_0-\ell_1>\varepsilon_{\mathrm{imp}}$
            \textbf{else} $-1$

        \If{$v=\text{adaptive}$ \textbf{and} $s=-1$}
            \State $\mathcal Q\gets\Call{Random}{p_0}$
        \Else
            \State $\mathcal Q\gets\Call{Directional}{p_0,su}$
            \If{$v=\text{hybrid}$}
                \State $\mathcal Q\gets\mathcal Q\cup\Call{Random}{p_0}$
            \EndIf
        \EndIf

        \State $p^\star\gets
            \Call{SimulateAndSelect}{\{p_0,p_1\}\cup\mathcal Q}$
        \State $z_0^{t_k}\gets
            \Call{WriteBack}{z_0^{t_k},p^\star}$
        \State $z\gets\Call{ReNoise}{z_0^{t_k},\epsilon,t_k}$
    \EndIf

    \State $z\gets\Call{SamplingStep}{z,\epsilon,t_k}$
\EndFor
\State $\widehat{\layout}\gets\Call{DecodeLayout}{z}$
\State Evaluate $\simloss
    (\widehat{\layout};\target,\context)$
\State \Return $\widehat{\layout}$
\end{algorithmic}
\end{algorithm}

\textsc{Directional} generates $m_{\mathrm d}$ candidates along the supplied direction using step size $h$, while \textsc{Random} generates $m_{\mathrm r}$ candidates using the configured random perturbations. \textsc{SimulateAndSelect} returns the candidate with the lowest simulator loss among those evaluated, including the baseline and probe.

\subsection{Geometric Perturbations}
\label{app:perturbations}

\begin{figure}[htbp]
    \centering
    \small
    \setlength{\tabcolsep}{0pt}
    \newlength{\stepframewidth}
    \setlength{\stepframewidth}{\dimexpr(\textwidth-39pt)/3\relax}
    \newcommand{\stepframe}[1]{%
        \raisebox{-0.5\height}{%
            \includegraphics[width=\stepframewidth]{#1}}}
    \begin{tabular}{@{}c@{\hspace{3pt}}c@{\hspace{3pt}}c@{\hspace{3pt}}c@{}}
        \makebox[30pt]{} & Frame 17 & Frame 18 & Frame 19 \\[3pt]
        \textbf{SRD}
        & \stepframe{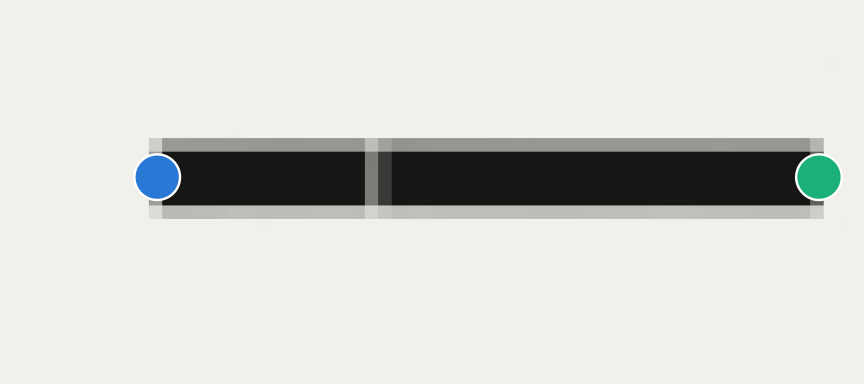}
        & \stepframe{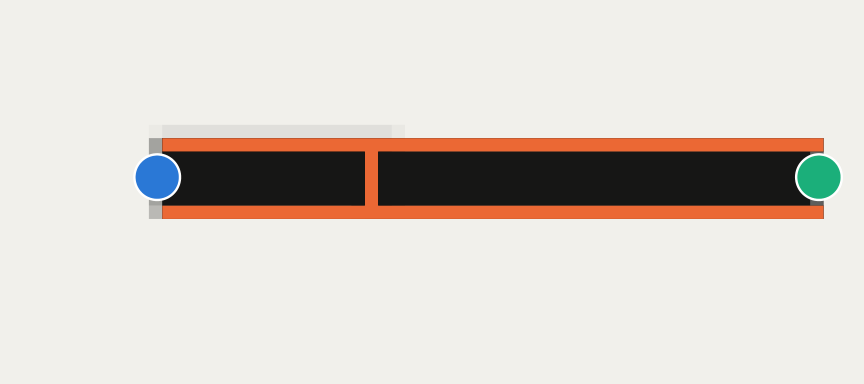}
        & \stepframe{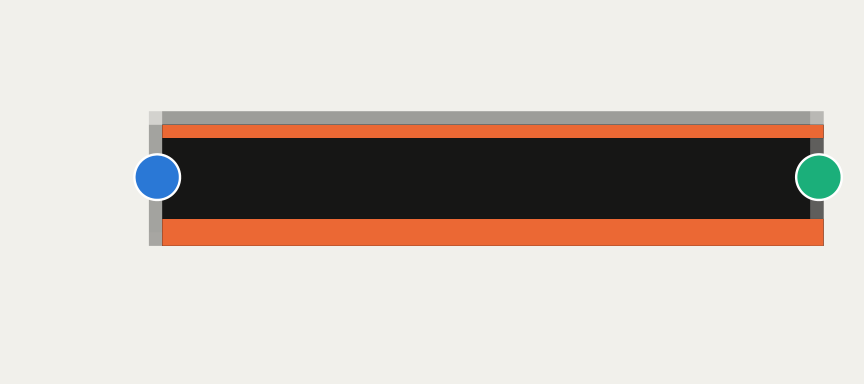} \\
        \noalign{\vskip4pt}
        CD
        & \stepframe{images/step17_shared.png}
        & \stepframe{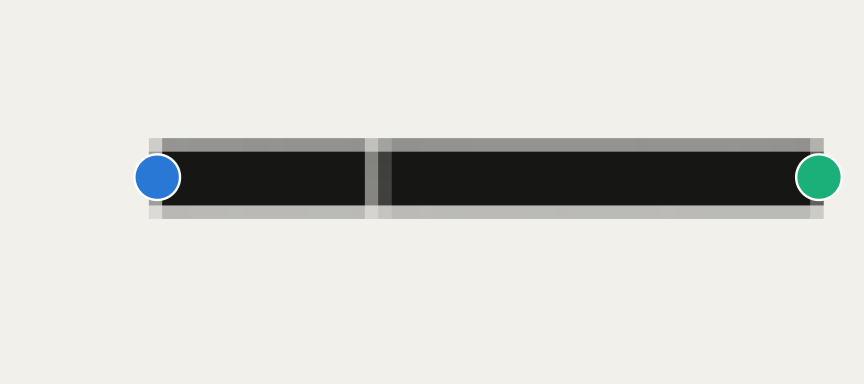}
        & \stepframe{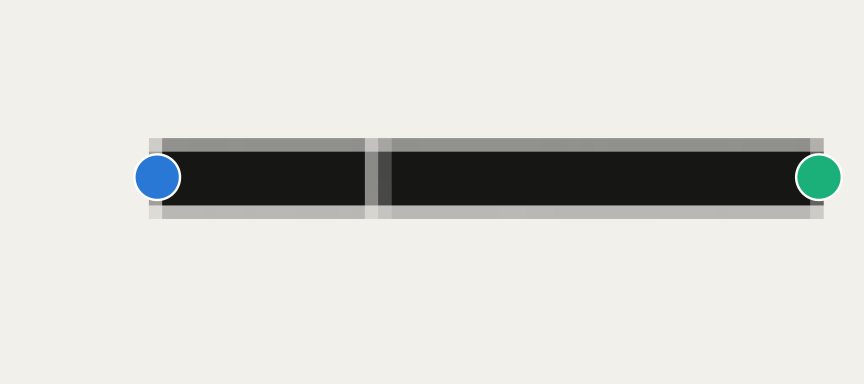}
    \end{tabular}
    \caption{Intermediate layouts at denoising steps 17–19 of a 20-step sampling process for SRD (top) and CD (bottom). Both methods start from the same Frame-17 layout. Black denotes metal. Orange marks the geometric perturbations applied by SRD. Blue and aqua dots mark ports 1 and 2.}
    \label{fig:per_step}
\end{figure}

SRD generates candidate geometries by perturbing the centers, widths, and heights of the extracted rectangles. Palace evaluates these candidates. SRD selects the candidate with the lowest simulator loss and writes its geometry back into the clean prediction. Figure~\ref{fig:per_step} shows denoising steps 17--19 of a 20-step sampling process. SRD and CD share the same layout at step~17. At steps~18 and~19, SRD adds metal along the existing boundaries and widens the conducting strip. The port locations remain fixed. In contrast, CD produces no visible change over these steps.

\subsection{Effect on the Diffusion Prior}
\label{app:prior_proximity}

\begin{figure}[htbp]
    \centering
    \includegraphics[width=0.45\textwidth]{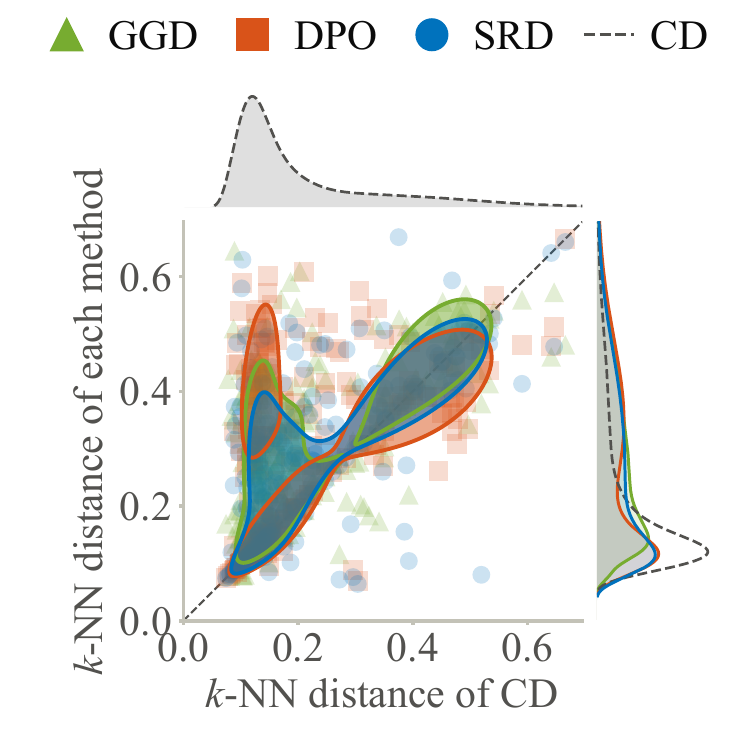}
    \caption{Nearest-neighbor distances ($k=1$) to the training set in DINOv3 embedding     space. Each point represents one sample. In the scatter plot, the horizontal axis reports the CD distance. The vertical axis reports the distance for the corresponding GGD, DPO, or SRD layout on the same target and seed.
    Colored contours show two-dimensional kernel density estimates (KDEs). Marginal KDE curves summarize the distance distributions along each axis.
    Points on the dashed diagonal indicate unchanged proximity to the training set; points above and below indicate layouts that are farther from and closer to it. This comparison shows how much each method moves layouts relative to the prior, and should be read alongside the losses in Table~\ref{tab:method_comparison}.}
    \label{fig:knn_distance}
\end{figure}

For each method, we compute the $L_2$ distance from each generated layout to its nearest training sample in DINOv3 embedding space~\citep{simeoni2025dinov3}. Figure~\ref{fig:knn_distance} compares these distances with those of CD. SRD retains distances close to CD for many samples while achieving the lowest mean and median S-parameter losses among the four compared methods (Table~\ref{tab:method_comparison}). These results suggest that SRD improves response accuracy with limited changes to the learned distribution.

% \FloatBarrier
% \Needspace{0.5\textheight}
\subsection{Qualitative Design Comparison}
\label{app:qualitative_comparison}

\begin{figure}[htbp]
    \centering
    \includegraphics[width=\textwidth]{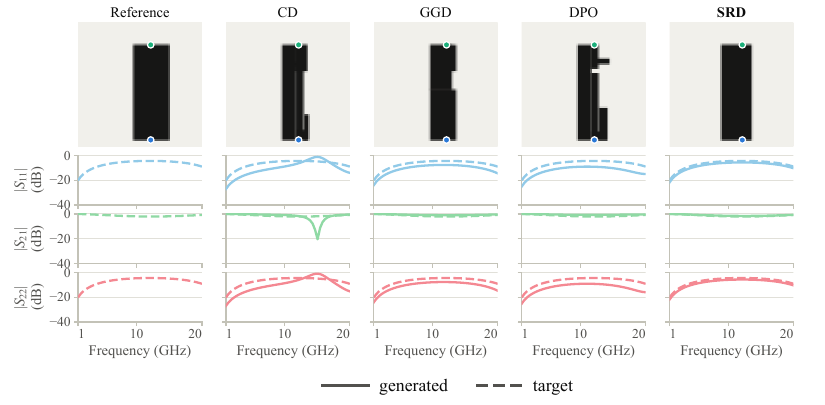}
    \setlength{\abovecaptionskip}{4pt}
    \caption{PCB layouts and their S-parameter magnitude responses. Solid and dashed curves indicate generated and target responses, respectively.}
    \label{fig:board_905_comparison}
\end{figure}

Figure~\ref{fig:board_905_comparison} compares an example target PCB with layouts generated by CD, GGD, DPO, and SRD and their S-parameter responses. CD exhibits a pronounced transmission dip near 15\,GHz that is absent from the target response. GGD and DPO remove this dip but retain substantial deviations in the reflection magnitudes. SRD more closely matches both the reflection and transmission responses across the frequency range, illustrating how geometric refinement improves electromagnetic agreement with the target.

\begin{figure}[htbp]
    \centering
    \begin{minipage}[t]{0.32\textwidth}
        \centering
        \small\textbf{SRD (adaptive)}\\[3pt]
        \includegraphics[width=\linewidth]{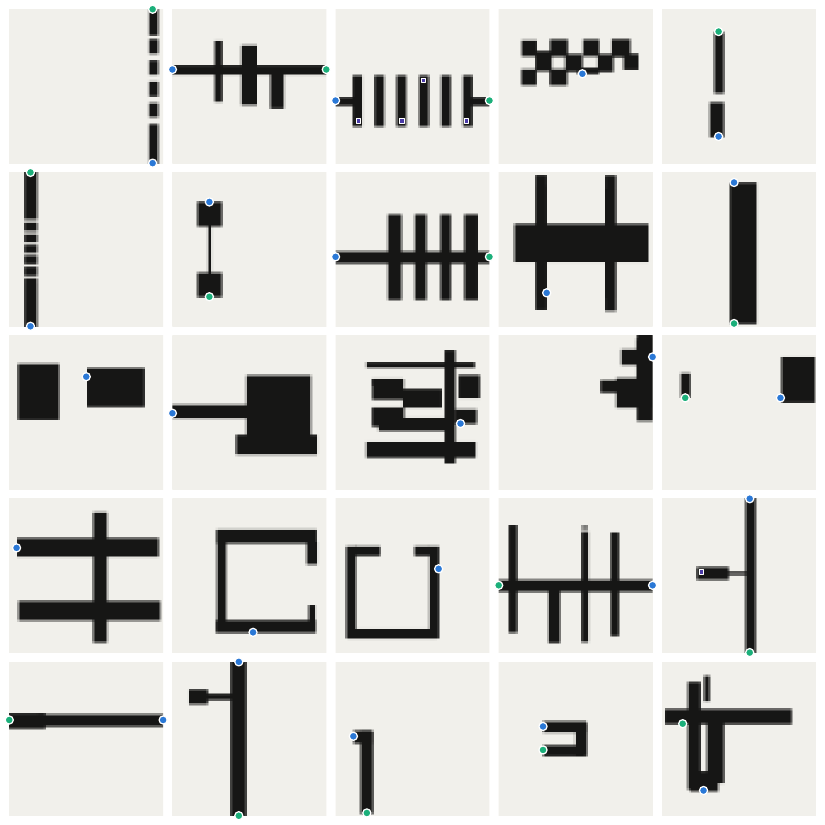}
    \end{minipage}\hfill
    \begin{minipage}[t]{0.32\textwidth}
        \centering
        \small\textbf{SRD (directional)}\\[3pt]
        \includegraphics[width=\linewidth]{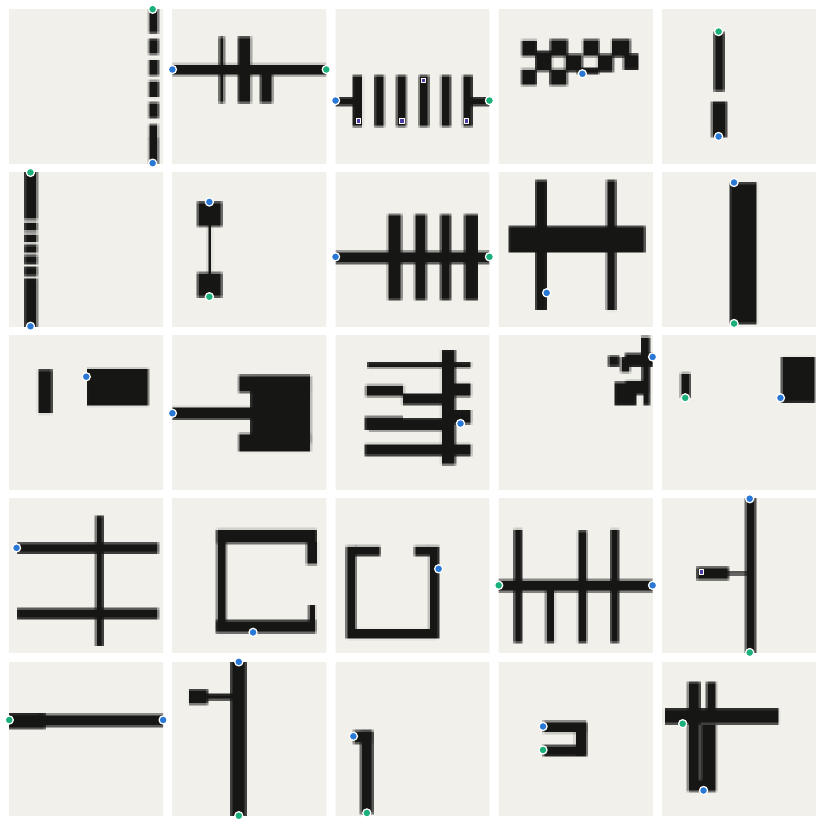}
    \end{minipage}\hfill
    \begin{minipage}[t]{0.32\textwidth}
        \centering
        \small\textbf{SRD (hybrid)}\\[3pt]
        \includegraphics[width=\linewidth]{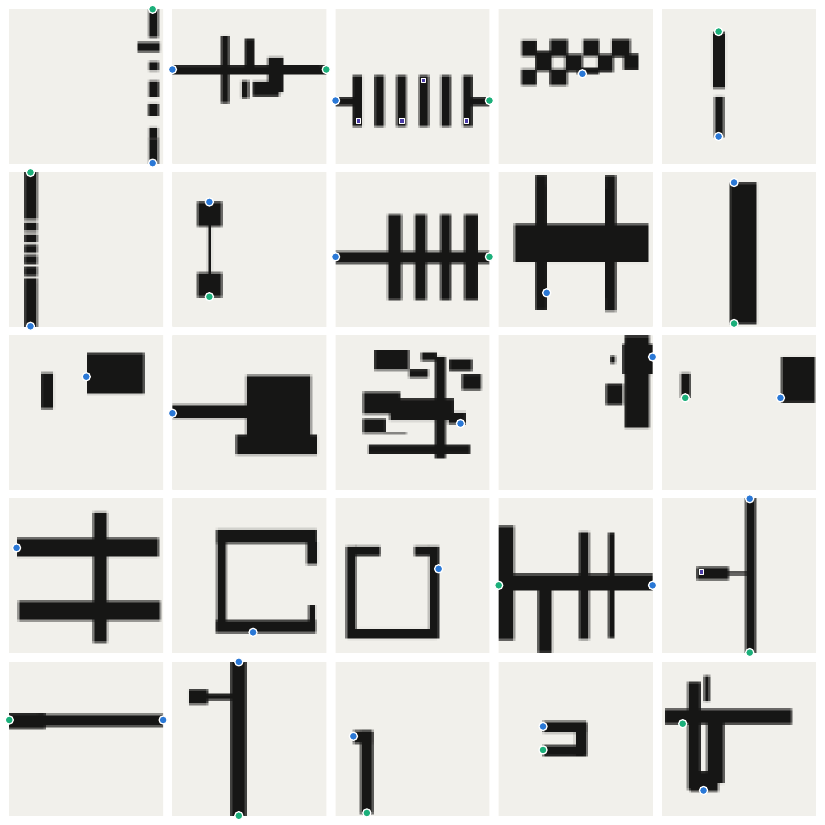}
    \end{minipage}
    \caption{Example PCB layouts generated by the three SRD variants. Each panel contains 25 layouts.}
    \label{fig:srd_variant_layouts}
\end{figure}

Figure~\ref{fig:srd_variant_layouts} shows example PCB layouts generated by the adaptive, directional, and hybrid SRD variants. In these examples, the adaptive and directional variants often produce similar overall structures, with differences in conductor widths and local branch geometry. The hybrid variant shows more pronounced geometric differences in several cases.

\FloatBarrier

\end{document}